\documentclass[a4paper,conference]{IEEEtran}

\usepackage{ifpdf}
\usepackage{cite}
\usepackage[pdftex]{graphicx}
\usepackage{array}
\usepackage{mdwmath}
\usepackage{mdwtab}
\usepackage{amssymb,latexsym}
\usepackage{stfloats}
\usepackage{amsmath}
\usepackage{subfig}

\DeclareRobustCommand*{\IEEEauthorrefmark}[1]{%
\raisebox{0pt}[0pt][0pt]{\textsuperscript{\footnotesize\ensuremath{#1}}}}

\begin{document}

\title{Skeleton-aware multi-scale heatmap regression for 2D hand pose estimation} 
\author{\IEEEauthorblockN{
Ikram Kourbane\IEEEauthorrefmark{1},
Yakup Genc\IEEEauthorrefmark{1}}
\IEEEauthorblockA{\IEEEauthorrefmark{1}
Gebze Technical University\\
Kocaeli, Turkey}
{\it {ikourbane,yahup.genc}@gtu.edu.tr}}
\maketitle

\begin{abstract}
Existing RGB-based 2D hand pose estimation methods learn the joint locations from a single resolution, which is not suitable for different hand sizes. To tackle this problem, we propose a new deep learning-based framework that consists of two main modules. The former presents a segmentation-based approach to detect the hand skeleton and localize the hand bounding box. The second module regresses the 2D joint locations through a multi-scale heatmap regression approach that exploits the predicted hand skeleton as a constraint to guide the model. Furthermore, we construct a new dataset that is suitable for both hand detection and pose estimation. We qualitatively and quantitatively validate our method on two datasets. Results demonstrate that the proposed method outperforms state-of-the-art and can recover the pose even in cluttered images and complex poses.

\end{abstract}


\IEEEpeerreviewmaketitle

\section{Introduction\label{section1}}

The hands are one of the most important and intuitive interaction tools for humans. Solving the hand pose estimation problem is crucial for many applications, including human-computer interaction, virtual reality and augmented reality. The earlier works in hand tracking use special hardware to track the hand, such as gloves and visual markers. Yet, these types of solutions are expensive and restrict the applications to limited scenarios. To this end, tracking hands without any device or markers is desirable. However, markerless hand trackers have their challenges due to articulations, self-occlusions, variations in shape, size, skin texture and color. 

The rapid development of deep learning techniques revolutionizes complex computer vision problems and outperforms conventional methods in many challenging tasks. Hand pose estimation is not an exception and deep convolution neural networks (CNNs) \cite{cnn} have been applied successfully in \cite{depth2,depth3,depth1}. These studies address the scenarios where the hand is tracked via an RGB-D camera. However, depth-enhanced data is not available everywhere, and they are expensive to utilize. Thus, estimating the hand pose from a single RGB image has been an active and challenging area of research, as they are cheaper and easier to use than RGB-D cameras \cite{zimmermann,spurr,mueller}.  

Recently deep learning-based methods achieve promising results. We can classify them into two broad categories: regression-based and detection-based. The regression approaches use CNNs as an automatic feature extractor to directly estimate the joint locations \cite{zimmermann, gomez,direct}. Although this approach is fast at inference time, direct regression is a difficult optimization problem. This is mainly due to its non-linear nature requiring many iterations and a lot of data for convergence. To overcome these limitations, recent works use probability density maps such as the heatmap to solve human and hand pose estimation problems \cite{mueller,heat,iqbal}. They divide the pose estimation problem into two steps. While the first one finds a dense pixel-wise prediction for each joint, the second step infers the joint locations by finding the maximum pixel in each heatmap. The heatmap representation helps the neural network to estimate the joint locations robustly and has a fast convergence property \cite{nodirect}. 

In this work, we focus on the problem of 2D hand pose estimation from a single RGB image. This task is also challenging due to the many degrees of freedom (DOF) and the self-similarity of the hand. The proposed approach has two principal components; The former estimates the hand skeleton to predict the bounding box using the well-know UNet architecture \cite{unet}. The second part presents a new multi-scale heatmap regression approach to estimate joint locations from multiple resolutions. Specifically, the network output is supervised on different scales to ensure accurate poses for different hand image sizes. This strategy helps the model for better learning of the contextual and the location information. Besides, our method uses the predicted hand skeleton as additional information to guide the network to robustly predict the 2D hand pose.

Furthermore, we create a new dataset suitable for both hand detection and 2D pose estimation tasks. This dataset includes bounding boxes, 2D keypoints, 3D pose and their corresponding temporal RGB images. We validate the proposed method on a common existing multiview hand pose dataset (LSMV). Furthermore, we extended our results to our newly created dataset (GTHD). Results demonstrate that our method generates accurate poses and outperforms three state-of-the-arts \cite{gomez, lie, kong}.

\section{Skeleton detection and bounding box localization} \label{detectionMethod}

\begin{figure}
\centering
\includegraphics[width=0.5\textwidth]{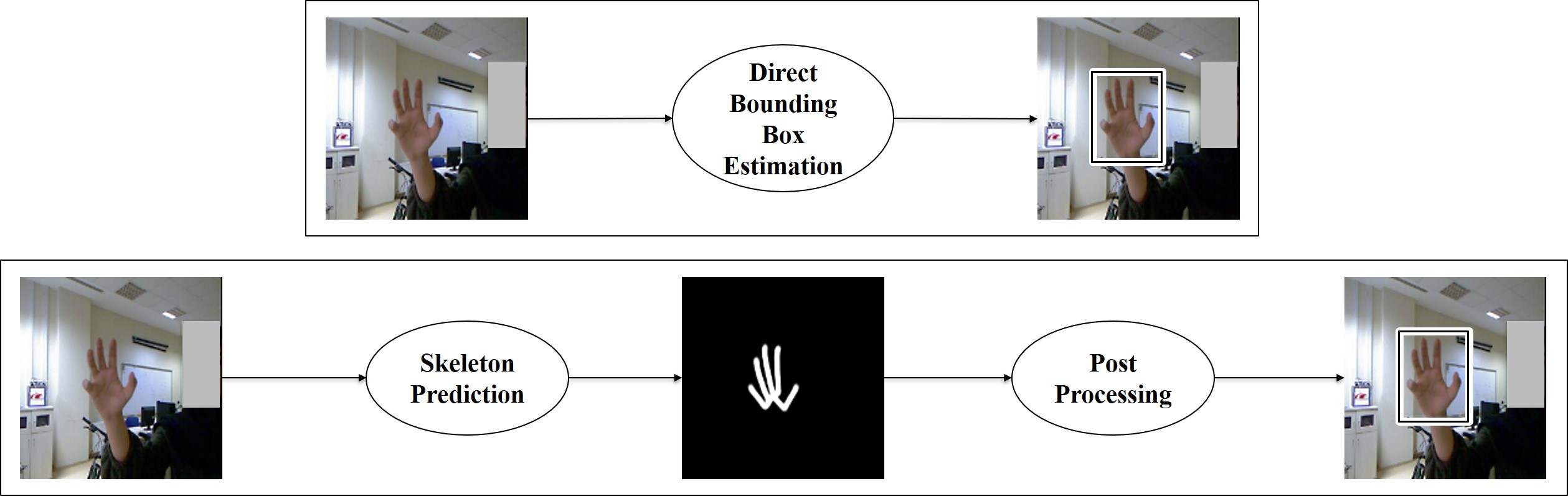}

\caption{The proposed method for hand bounding box detection. Unlike many deep learning approaches that use Faster R-CNN \cite{fasterRCNN} model to directly estimate the bounding box (top), we predict the skeleton image and infer the bounding box in a post-processing step (bottom).}

 \label{detection}
\end{figure}

We represent the detected hand location in an image by a rectangular region with four corners. Faster-RCNN \cite{fasterRCNN} type of deep network models directly regress the four corner coordinates from the given hand image. Alternatively, we can predict the 2D hand skeleton and extract the bounding box in a post-processing step (Figure~\ref{detection}). Direct regression of the bounding box is useful for hand cropping but cannot be further exploited for other tasks. In contrast, estimating the hand skeleton includes useful information that guides the 2D pose estimation. Also, the segmentation task is less challenging than predicting the bounding box. Of course, one needs to have the training data with corresponding skeletons. We can obtain this type of data using a 3D hand tracker and an RGB camera to provide the 2D key-points (see Section~\ref{dataset}). We create the ground truth data for the skeleton by connecting the joints in each finger. Also, we attach the palm to the ends of each finger. Additionally, we represent each joint location by the standardized Gaussian blob.

We can treat hand skeleton data as a segmentation mask. Thus, we use the UNet architecture \cite{unet} since it is one of the best encoder-decoder architectures for semantic segmentation. It has two major properties. The first one is the skip connections between the encoder and the decoder layers that enable the network to learn the location and the contextual information. The second property is its symmetry, leading to better information transfer and performances. The model outputs single feature maps on which we apply a \textit{sigmoid} activation function to bound the prediction values between 0 (background) and 1 (hand). We localize the bounding box using a post-processing step, in which we identify the foreground pixels, and then we apply a region growing algorithm. In our case, the horizontal and vertical boundaries of the recovered regions are reported as the location of the detected hand. Our model robustly differentiates between the skin of the hand and that of the face. Also, it can detect the hand even in cluttered images or different lightning conditions (see Section~\ref{detectionResults}).


We trained the model for 20 epochs using a batch size of 8. Concerning the loss function, we did two experiments. In the first one, we only used the \textit{$L_1$} loss function, which can not robustly localize the skeleton and adversely affecting the bounding box localization results. On the other hand, using the combination of \textit{$L_1$} loss and a \textit{SoftDice} loss with their empirical weights can robustly localize the hand.

\begin{equation}
L_1(x, \hat{x}) =\|\ x - \hat{x} \| ^{1}_{1}
\end{equation}
\begin{equation}
SoftDice(x, \hat{x}) = 1 - \frac{2\hat{x}^T x}{\|\hat{x}\| ^{2}_{2} + \|x\| ^{2}_{2}}
\end{equation}
\begin{equation}
Total(x, \hat{x}) = \lambda_1 L_1(x, \hat{x}) + \lambda_2 SoftDice(x, \hat{x})
\end{equation}

Where: $x$, $\hat{x}$, $\lambda_1$ and $\lambda_2$ represent the ground truth skeleton, the predicted skeleton and the two hyperparameters of the loss function (set to 0.4 and 0.6, respectively).

\section{Multi-scale heatmaps regression} \label{poseMethod}
Most of the existing hand pose estimation methods predict the heatmaps at a single-scale. However, the hand in the original image can have several sizes (close/far hands). Hence, when we use a single scale image, the cropped hand image size cannot be suitable for all the dataset samples. To address this limitation, we propose a multi-scale heatmaps regression architecture that performs the back-propagation process for many resolutions providing better joint learning for both large and small hands. Moreover, the cropped hand image would include some parts of the background. To overcome this problem, we employ the predicted hand skeleton to act as an attention mechanism for the network to focus on hand pixels. This makes the 2D pose regression task less challenging to optimize. 
\begin{figure}
\centering

\includegraphics[width=0.5\textwidth]{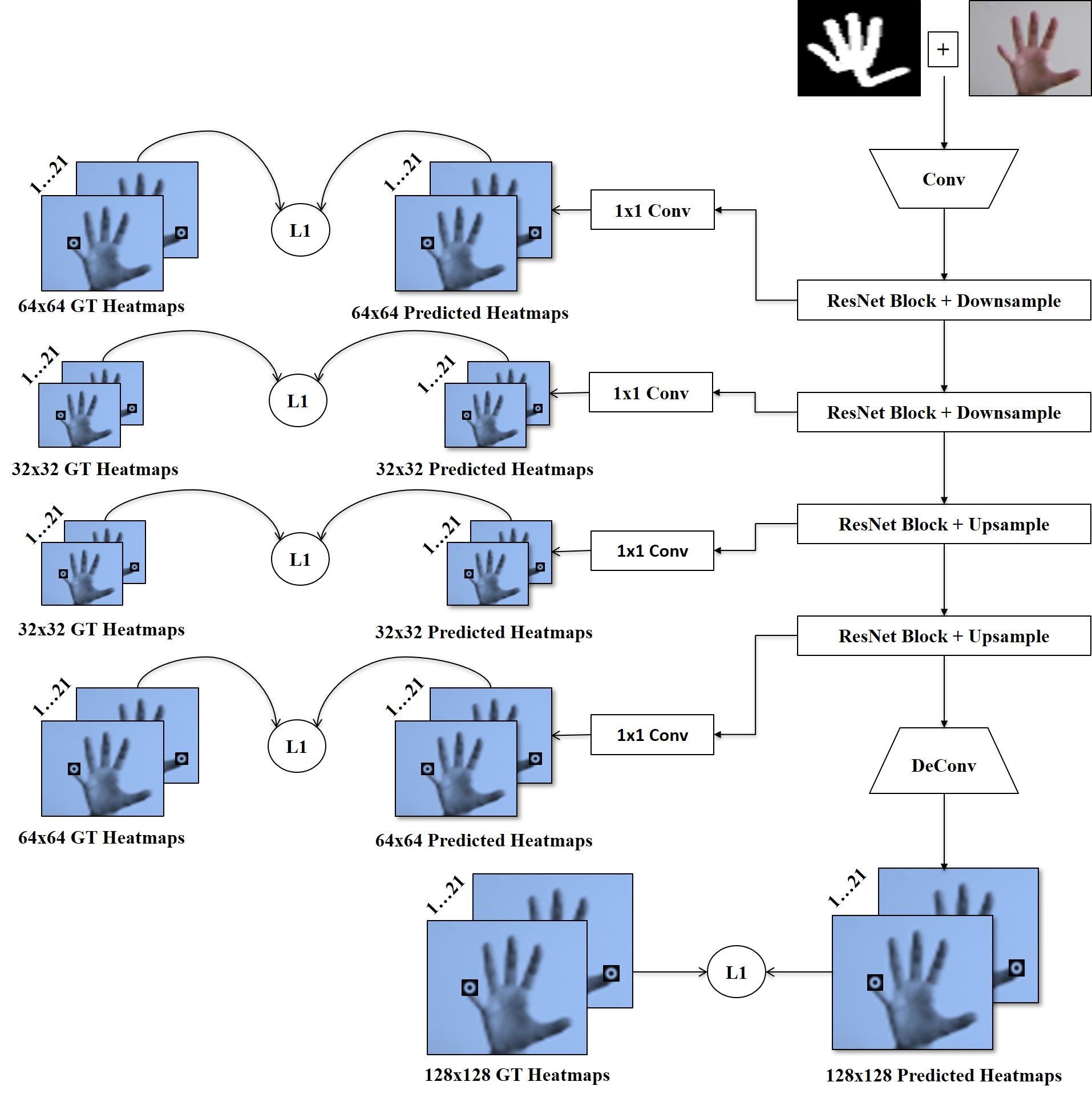}
\caption{The overall architecture of the proposed 2D hand pose estimation approach uses the hand skeleton as a constraint and estimates the joint heatmaps from multiple scales.}
\label{multiscale}
\end{figure}

Figure~\ref{multiscale} shows our skeleton-aware multi-scale heatmaps network approach for 2D hand pose estimation. We feed the concatenation of the cropped hand image and the predicted skeleton to the first convolution layer. The latter is followed by two downsampling ResNet blocks, two upsampling ResNet blocks, and a final transposed convolution layer that recovers the input resolution. After each downsampling (similarly upsampling), we apply a $1\times1$ convolution layer followed by a \textit{sigmoid} activation function to output 21 or 20 feature maps representing the heatmaps in GTHD or LSMV datasets, respectively. The heatmaps resolution is divided/multiplied by two after each downsampling/upsampling. In test time, we calculate a weighted average of the predicted five heatmaps to find the coordinate of the 2D keypoints. We formulate the loss function as:

\begin{equation}
L(x, \hat{x}) = \sum_{i=1}^{k} \delta_i{} \|\ x_{i} - \hat{x}_{i} \| ^{2}_{2}
\end{equation}

Where: $k$ is the number of scales including the full resolution output, and $\delta_{i}$ is the weight given for each scale. In our experiments we choose $k=5$ and $\delta_{i}$ is set be 1, 1/2 and 1/4 for scales 128, 64 and 32 respectively.

\section{Experiments}
 \label{experiments}

\subsection{Implementation details}
 \label{dataset}

Our method has been implemented and tested on two different datasets. The first one is LSMV \cite{gomez}, which is one of the large-scale datasets that provide the hand bounding boxes, the 2D key-points as well as the 3D pose. We split the data into 60000, 15000, and 12760 samples for the training set, validation set, and test set, respectively. 

\begin{figure}
\centering
\includegraphics[width=0.48\textwidth]{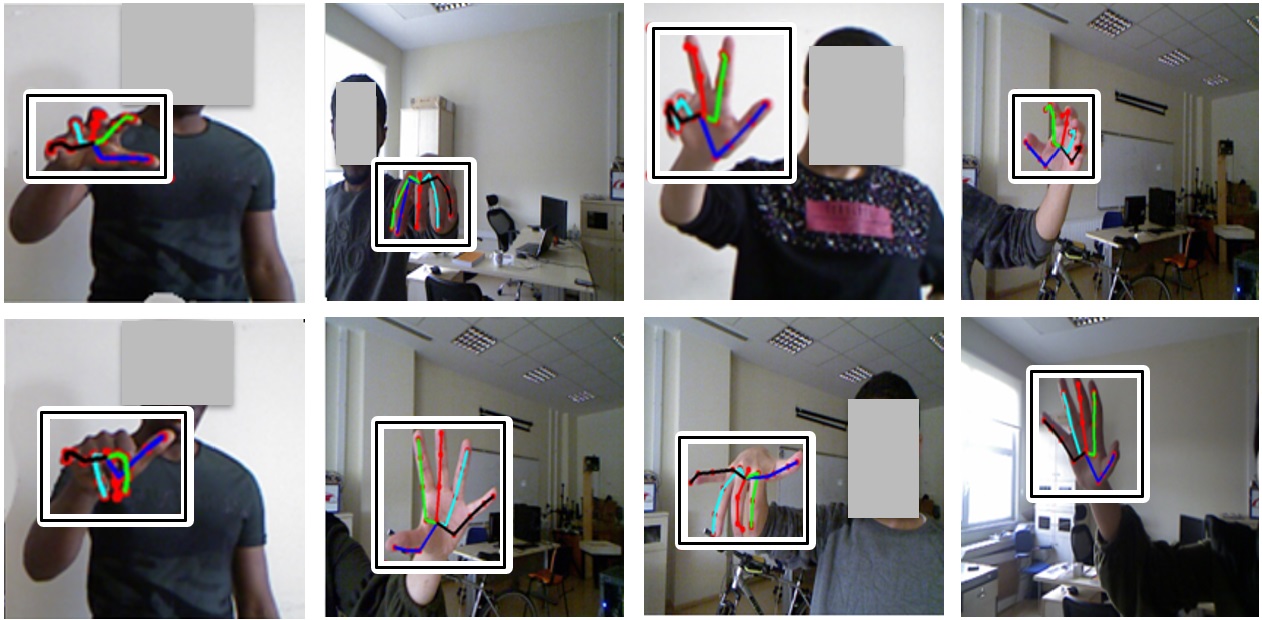}
\caption{Examples of our hand dataset images having the bounding boxes and 21 joints annotations taken from four subjects and covering many pose and backgrounds.}
\label{samples}
\end{figure}

To train both the hand detector and the hand pose estimator, we have built our own dataset (GTHD) using an RGB camera and a Leap Motion sensor \cite{leap}. The RGB camera provides an image with a resolution of $640\times480$ pixels. The leap motion controller is a combination of hardware and software that senses the fingers of the hand to provide the 3D joint locations. Hence, a projection process from 3D space to the 2D image plane is necessary. We achieve this goal in two steps. In the first one, we use OpenCV to estimate specific intrinsic parameters of the camera. In the second step, we estimate the extrinsic parameters between the leap motion controller and the camera. To get the correct pose with its corresponding image, we synchronize the two sensors in time. Finally, to find the rotation and translation matrices, we manually mark one key-point in a set of hand images and solve the \textit{PnP} problem by computing the 3D-2D correspondences \cite{ziss}. Figure~\ref{samples} illustrates the results of the calibration process. We randomly split the GTHD dataset into a training set ($75\%$), a validation set ($10\%$) and a test set ($15\%$).

We report the performance of that module using \textit{Accuracy}, \textit{Precision}, \textit{Recall} and \textit{F1}. Also, we calculate the Area Under ROC Curve ( \textit{AUC}) for GTHD datasets since it measures how well the two classes (Hand and NoHand) are separable. It calculates the trade-off between the true positive rate and the false positive rate. Furthermore, we report the Intersection over Union ( \textit{IOU}) metric to quantify our model performance in hand bounding box detection task. It evaluates the predicted bounding boxes by comparing them against the ground truth.

To quantitatively evaluate the performance of the proposed 2D hand pose regression methods, we use the Probability of Correct Keypoint (PCK) metric \cite{pck} as it is used frequently in human and hand pose estimation tasks. It considers the predicted joints as correct if the distance to the ground truth joint is within a certain threshold. We used a normalized threshold by dividing all the joints values by the size of the hand bounding box. Also, for additional quantification of the performance of the proposed method, we report the low joint pixel error (MJPE) per keypoint over the input hand image with $128\times128$ resolution.

We train the models for 30 epochs using a batch size of 8 and Adam as an optimizer. We initialize the learning rate to $0.01$, and we decrease it after every eight epochs by $10\%$. We conduct all experiments on NVIDIA GTX 1080 GPU using PyTorch v1.6.0. 

\begin{figure*}[t]
\centering
\includegraphics[width=0.99\textwidth]{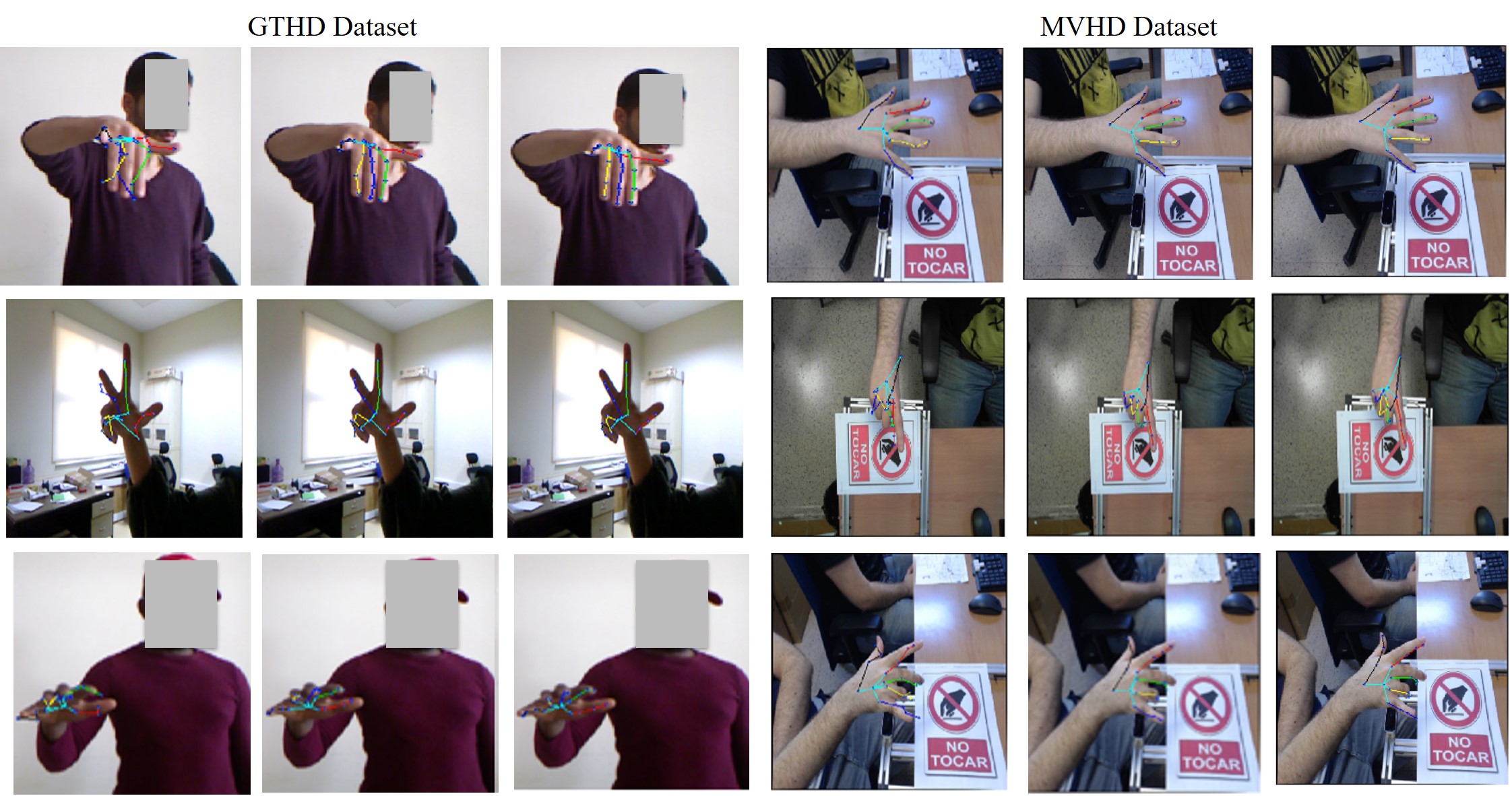}
\caption{Qualitative results on GTHD and LSMV datasets. The columns from left to right in each image show: the direct regression proposed in \cite{gomez}, our proposed skeleton aware multi-scale heatmaps regression and the ground truth joints.}
\label{pose}
\end{figure*}

\subsection{Hand detection }
 \label{detectionResults}
\begin{figure}
\centering
\includegraphics[width=0.5\textwidth]{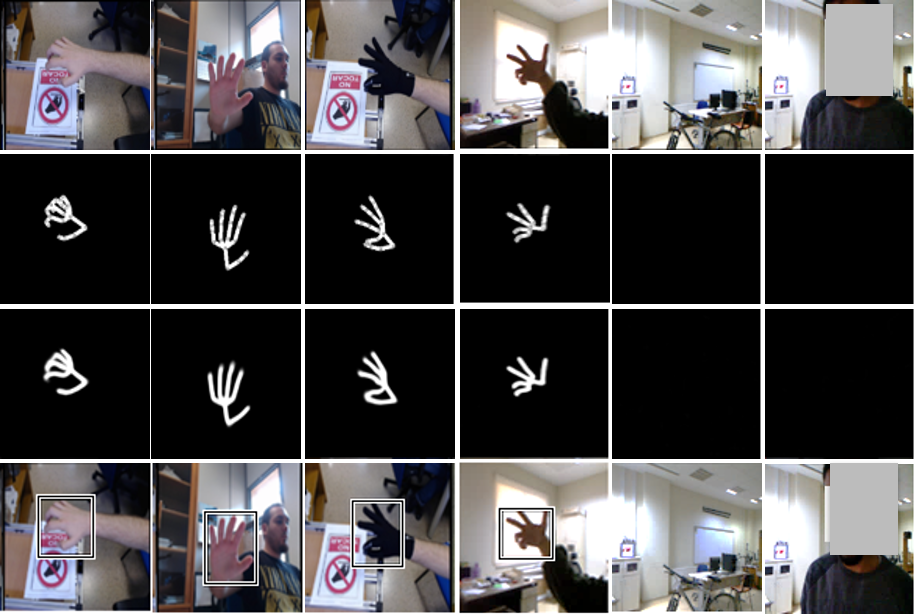}
\caption{The results of the skeleton estimation and the bounding box localization on LSMV and GTHD datasets. The rows from top to down show: the input image, the ground truth skeleton, the predicted skeleton, and the obtained bounding boxes.}
\label{multiview}
\end{figure}

Our approach can robustly estimate the hand skeleton and localize the bounding box for LSMV and GTHD datasets. Furthermore, it does not produce any false positives for background images or images with people who do not show their hands (see Figure~\ref{multiview}). 

\begin{figure}
\centering
\includegraphics[width=0.49\textwidth]{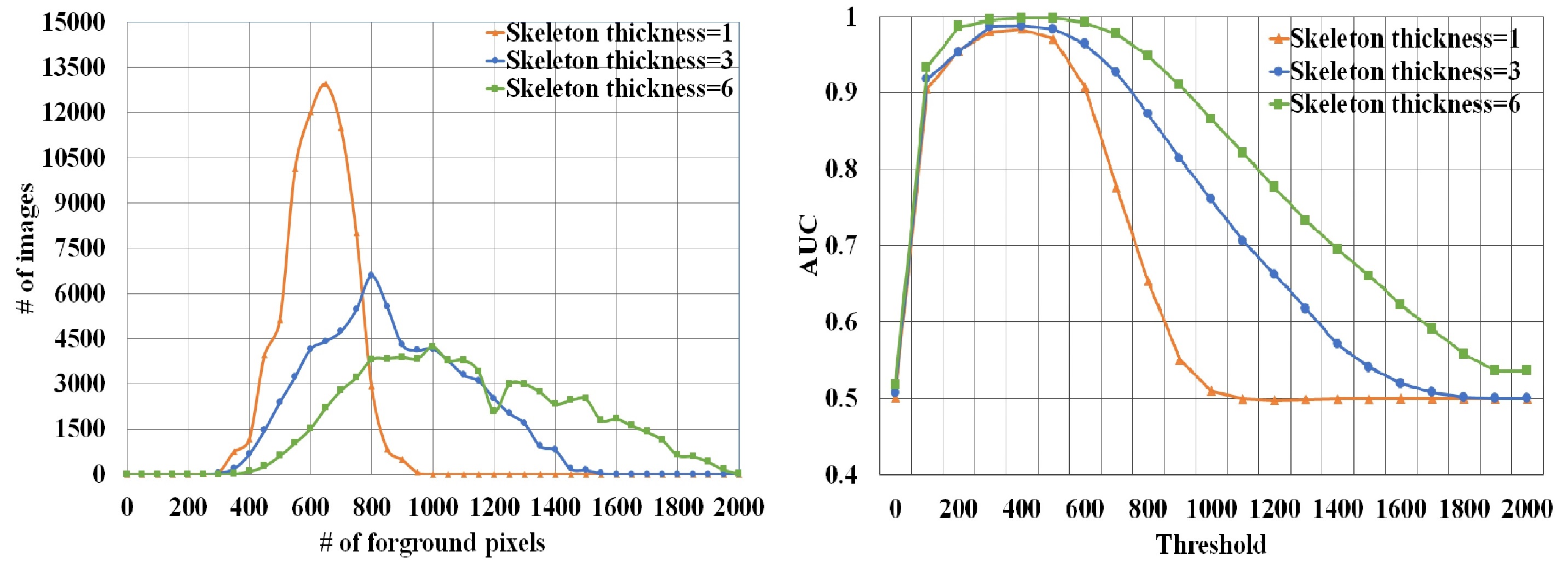}
\caption{The impact of the threshold selection on the AUC performances.}
\label{threshold}
\end{figure}

The correct threshold for selecting \textit{Hand} from \textit{NoHand} depends on the data. A robust threshold should eliminate the noise and be in an interval that does not miss samples from the dataset distribution. In other words, the selected threshold should decrease both the false-negative rate (adding samples from the \textit{NoHand} class) and false positive rate (missing samples from the \textit{Hand} class) to achieve high performance and robustly detect the hand. Figure~\ref{threshold}. shows that selecting a threshold from the interval $[200,400]$ is the best choice for our dataset. Also, the thickest skeleton representation seems to be more robust to the noise. It outperforms the other representations and achieves a higher performance ($AUC=0.99$). Finally, our approach records high scores of $Accuracy=0.99$, $Precision=0.97$, $Recall =0.99$ and $F1=0.98$.

We do not report the AUC for the MVHD dataset because it does not have images without hands. Nevertheless, we predict the skeleton representation to extract the hand bounding boxes and perform our proposed 2D hand pose estimation method. Also, we report \textit{IOU} in Table~\ref{iou} showing that the proposed method outperforms Faster-RCNN \cite{fasterRCNN}.  

\begin{table}
\centering
\caption{Bounding box evaluation on MVHD and our GTHD dataset with IOU. 
}
\begin{tabular}{|c|c|c|}
\hline
Dataset &  Faster-RCNN \cite{fasterRCNN} & Ours\\
\hline
GTHD & 0912 & 0.923\\
\hline
LSMV  &0.895  &  0.917\\
\hline

\end{tabular}
\label{iou}
\end{table}

\subsection{Pose estimation}
 \label{experiments}

We compare the proposed pose estimation approach against three-deep learning-based methods \cite{gomez, lie, kong}. Before giving the RGB image as input to the model, we resize and normalize the datasets by subtracting the mean from all the images. The number of heatmaps is the number of joints where we represent each one with a Gaussian blob in a map of the same size of the image. The coordinate of the joint is the location of the highest value in the heatmap. We find them by applying the \textit{argmax} function. 

Our baseline is \cite{gomez} that uses ResNet-50 architecture \cite{resnet} to directly regress the 2D joints from RGB images. The other deep-based methods \cite{lie, kong} are two of the existing state-of-the-art in 2D hand pose estimation. Moreover, to demonstrate the effectiveness of the proposed approach, we conduct two additional experiments. The former applies single-scale heatmap regression using UNet architecture \cite {unet} on $128\times128$ resolution images. The second experiment performs our multi-scale heatmaps regression without the skeleton information.

\begin{figure}
\centering
\includegraphics[width=0.5\textwidth]{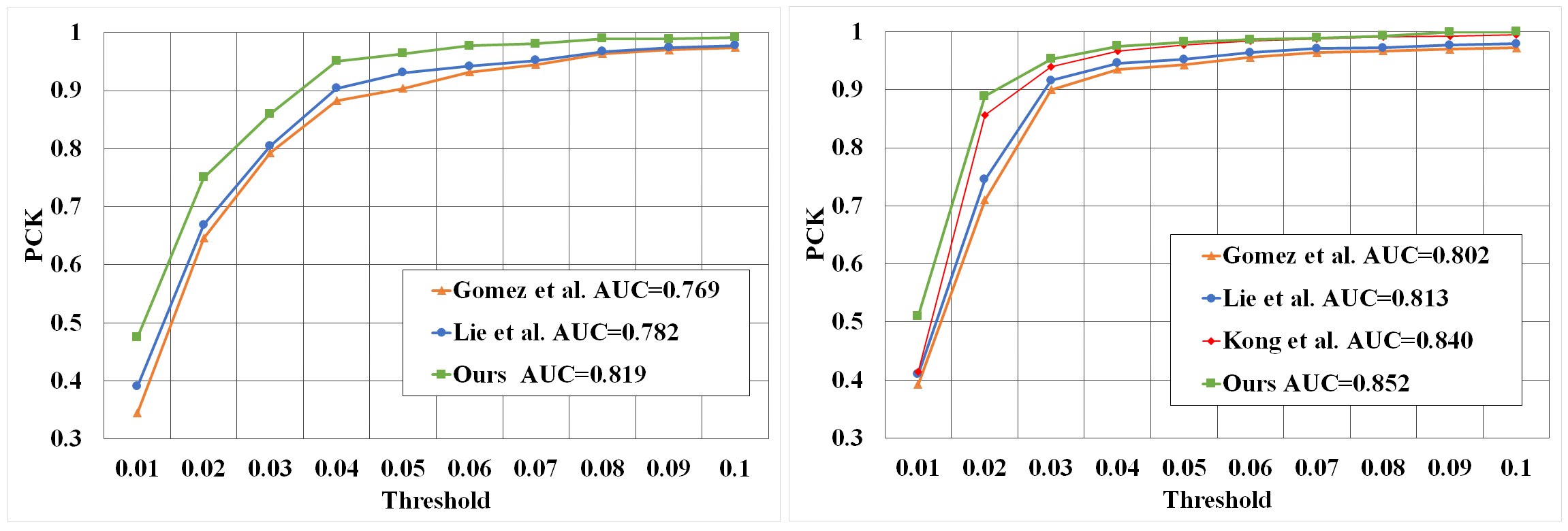}
\caption{Quantitative comparison of the proposed 2D hand pose estimation with the other methods \cite{gomez,lie,kong} using PCK metric. Left for GTHD and right for LSMV.}
\label{pck}
\end{figure}

Figure~\ref{pose} shows some randomly selected test images on both datasets. The proposed method can robustly estimate the 2D hand pose even in complex poses and cluttered images.  Furthermore, the proposed skeleton-aware multi-scale heatmaps regression method outperforms the state-of-the-art \cite{gomez, lie, kong}. It achieves a high PCK score (0.98) with a small threshold on the two datasets (Figure~\ref{pck}). Besides, our proposed method for 2D hand pose estimation provides more improvement for our dataset since it has more complex poses, face occlusion cases, and lighting condition variations (Figure~\ref{pck} and Table~\ref{2D}). 

The hand skeleton representation improves the proposed multi-scale heatmaps regression method since it constrains the 2D pose estimation task. Also, it favorably performs against the single-scale heatmaps regression method. (Table~\ref{2D}).

\begin{table}
\centering
\caption{Comparison with the state-of-the-art methods on GTHD and LSMV datasets with Mean pixel errors.}
\begin{tabular}{|c|c|c|}
\hline
Methods &GTHD  & LSMV \\
\hline
Gomez et al \cite{gomez} & 13.20 & 10.00 \\
\hline
Lie et al. \cite{lie} & 6.25 &  8.05\\
\hline
Single-scale heatmaps regression using \cite{unet} & 7.33 & 5.87\\
\hline
Ours w/o skeleton & 5.89 &  4.95\\
\hline
Ours  &\textbf{5.51} & \textbf{4.67}\\
\hline
\end{tabular}
\label{2D}
\end{table}

\section{Conclusion}
In this work, we propose a new learning-based method for 2D hand pose estimation. It performs multi-scale heatmaps regression and uses the hand skeleton as additional information to constrain the regression problem. It provides better results compared with the direct regression and single-scale heatmaps regression. Also, we present a new method for hand bounding box localization that first estimate the hand skeleton and then extract the bounding box. This approach provides accurate results since it learns more information from the skeleton. Furthermore, we introduce a new RGB hand pose dataset that can use both for hand detection and 2D pose estimation tasks. For future work, we plan to exploit our 2D hand pose estimation method to improve the 3D hand pose estimation from an RGB image. Also, we plan to incorporate other constraints that can restrict the hand pose estimation problem.
 
\bibliographystyle{ieeetr} 
\bibliography{reference}

\begin{thebibliography}{10}

\bibitem{cnn}
A.~Krizhevsky, I.~Sutskever, and G.~E. Hinton, ``Imagenet classification with
  deep convolutional neural networks,'' {\em Communications of the ACM},
  vol.~60, no.~6, pp.~84--90, 2017.

\bibitem{depth2}
Y.~Yang, C.~Feng, Y.~Shen, and D.~Tian, ``Foldingnet: Point cloud auto-encoder
  via deep grid deformation,'' in {\em Proceedings of the IEEE Conference on
  Computer Vision and Pattern Recognition}, pp.~206--215, 2018.

\bibitem{depth3}
S.~Li and D.~Lee, ``Point-to-pose voting based hand pose estimation using
  residual permutation equivariant layer,'' in {\em Proceedings of the IEEE
  Conference on Computer Vision and Pattern Recognition}, pp.~11927--11936,
  2019.

\bibitem{depth1}
C.~Wan, T.~Probst, L.~Van~Gool, and A.~Yao, ``Crossing nets: Combining gans and
  vaes with a shared latent space for hand pose estimation,'' in {\em
  Proceedings of the IEEE Conference on Computer Vision and Pattern
  Recognition}, pp.~680--689, 2017.

\bibitem{zimmermann}
C.~Zimmermann and T.~Brox, ``Learning to estimate 3d hand pose from single rgb
  images,'' in {\em Proceedings of the IEEE international conference on
  computer vision}, pp.~4903--4911, 2017.

\bibitem{spurr}
A.~Spurr, J.~Song, S.~Park, and O.~Hilliges, ``Cross-modal deep variational
  hand pose estimation,'' in {\em Proceedings of the IEEE Conference on
  Computer Vision and Pattern Recognition}, pp.~89--98, 2018.

\bibitem{mueller}
F.~Mueller, F.~Bernard, O.~Sotnychenko, D.~Mehta, S.~Sridhar, D.~Casas, and
  C.~Theobalt, ``Ganerated hands for real-time 3d hand tracking from monocular
  rgb,'' in {\em Proceedings of the IEEE Conference on Computer Vision and
  Pattern Recognition}, pp.~49--59, 2018.

\bibitem{gomez}
F.~Gomez-Donoso, S.~Orts-Escolano, and M.~Cazorla, ``Large-scale multiview 3d
  hand pose dataset,'' {\em Image and Vision Computing}, vol.~81, pp.~25--33,
  2019.

\bibitem{direct}
J.~Carreira, P.~Agrawal, K.~Fragkiadaki, and J.~Malik, ``Human pose estimation
  with iterative error feedback,'' in {\em Proceedings of the IEEE conference
  on computer vision and pattern recognition}, pp.~4733--4742, 2016.

\bibitem{heat}
G.~Papandreou, T.~Zhu, N.~Kanazawa, A.~Toshev, J.~Tompson, C.~Bregler, and
  K.~Murphy, ``Towards accurate multi-person pose estimation in the wild,'' in
  {\em Proceedings of the IEEE Conference on Computer Vision and Pattern
  Recognition}, pp.~4903--4911, 2017.

\bibitem{iqbal}
U.~Iqbal, P.~Molchanov, T.~Breuel Juergen~Gall, and J.~Kautz, ``Hand pose
  estimation via latent 2.5 d heatmap regression,'' in {\em Proceedings of the
  European Conference on Computer Vision (ECCV)}, pp.~118--134, 2018.

\bibitem{nodirect}
A.~Bulat and G.~Tzimiropoulos, ``Human pose estimation via convolutional part
  heatmap regression,'' in {\em European Conference on Computer Vision},
  pp.~717--732, Springer, 2016.

\bibitem{unet}
O.~Ronneberger, P.~Fischer, and T.~Brox, ``U-net: Convolutional networks for
  biomedical image segmentation,'' in {\em International Conference on Medical
  image computing and computer-assisted intervention}, pp.~234--241, Springer,
  2015.

\bibitem{lie}
S.~Li and A.~B. Chan, ``3d human pose estimation from monocular images with
  deep convolutional neural network,'' in {\em Asian Conference on Computer
  Vision}, pp.~332--347, Springer, 2014.

\bibitem{kong}
D.~Kong, Y.~Chen, H.~Ma, X.~Yan, and X.~Xie, ``Adaptive graphical model network
  for 2d handpose estimation,'' {\em In Proceedings of the British Machine
  Vision Conference (BMVC)}, 2019.

\bibitem{fasterRCNN}
S.~Ren, K.~He, R.~Girshick, and J.~Sun, ``Faster r-cnn: Towards real-time
  object detection with region proposal networks,'' {\em IEEE transactions on
  pattern analysis and machine intelligence}, vol.~39, no.~6, pp.~1137--1149,
  2016.

\bibitem{leap}
L.~E. Potter, J.~Araullo, and L.~Carter, ``The leap motion controller: a view
  on sign language,'' in {\em Proceedings of the 25th Australian computer-human
  interaction conference: augmentation, application, innovation,
  collaboration}, pp.~175--178, 2013.

\bibitem{ziss}
P.~Beardsley, D.~Murray, and A.~Zisserman, ``Camera calibration using multiple
  images,'' in {\em European Conference on Computer Vision}, pp.~312--320,
  Springer, 1992.

\bibitem{pck}
T.~Simon, H.~Joo, I.~Matthews, and Y.~Sheikh, ``Hand keypoint detection in
  single images using multiview bootstrapping,'' in {\em Proceedings of the
  IEEE conference on Computer Vision and Pattern Recognition}, pp.~1145--1153,
  2017.

\bibitem{resnet}
K.~He, X.~Zhang, S.~Ren, and J.~Sun, ``Deep residual learning for image
  recognition,'' in {\em Proceedings of the IEEE conference on computer vision
  and pattern recognition}, pp.~770--778, 2016.

\end{thebibliography}

\end{document}